\pdfoutput=1
\documentclass[11pt]{article}

\usepackage{emnlp2021}
\usepackage{times}
\usepackage{latexsym}
\usepackage[T1]{fontenc}
\usepackage[utf8]{inputenc}

\usepackage{microtype}
\usepackage{multirow}
\usepackage{graphicx}  %Required
\usepackage{subcaption}
\usepackage{adjustbox}
\usepackage{amsmath}

\title{Learning to Share by Masking the Non-shared for Multi-domain Sentiment Classification}

\author{Jianhua Yuan, Yanyan Zhao, Bing Qin, Ting Liu \\
	Research Center for Social Computing and Information Retrieval, \\
	Faculty of Computing, \\
	Harbin Institute of Technology, Harbin	150001, China \\
	\texttt{ \{jhyuan, yyzhao, qinb, tliu\}@ir.hit.edu.cn} 
}

\begin{document}
\maketitle
\begin{abstract}
Multi-domain sentiment classification deals with the scenario where labeled data exists for multiple domains but insufficient for training effective sentiment classifiers that work across domains.  Thus, fully exploiting sentiment knowledge shared across domains is crucial for real world applications. While many existing works try to extract domain-invariant features in high-dimensional space, such models fail to explicitly distinguish between shared and private features at text-level, which to some extent lacks interpretablity.  Based on the assumption that removing domain-related tokens from texts would help improve their domain-invariance,  we instead first transform original sentences to be \textit{domain-agnostic}. To this end, we propose the BertMasker network which explicitly masks domain-related words from texts, learns domain-invariant sentiment features from these domain-agnostic texts, and uses those masked words to form domain-aware sentence representations.  Empirical experiments on a well-adopted multiple domain sentiment classification dataset demonstrate the effectiveness of our proposed model on both multi-domain sentiment classification and cross-domain settings, by increasing the accuracy by 0.94\% and 1.8\% respectively. Further analysis on masking proves that removing those domain-related and sentiment irrelevant tokens decreases  texts' domain distinction, resulting in the performance degradation of a BERT-based domain classifier by over 12\%.
\end{abstract}

\section{Introduction}
Sentiment classification \cite{pang-etal-2002-thumbs,liu2012sentiment} is one of the key tasks in Natural Language Processing. Recent success of sentiment classification relies heavily on deep neural networks trained with a large number of carefully annotated data. However, as the diversity of domains leads to the discrepancy of sentiment features, models trained on existing domains may not perform ideally on the domain of interest. Meanwhile, as not all domains have adequate labeled data,  it is necessary to leverage existing annotations from multiple domains. For instance, in both DVD and Video domains, \textit{picture} and \textit{animation} can be opinion targets and \textit{thrilling} and \textit{romantic} are frequent polarity words.   Exploiting such sharedness would help improve both in-domain and out-of-domain sentiment classification results.
	
\begin{figure}
	\centering
	\includegraphics[width=3.1in]{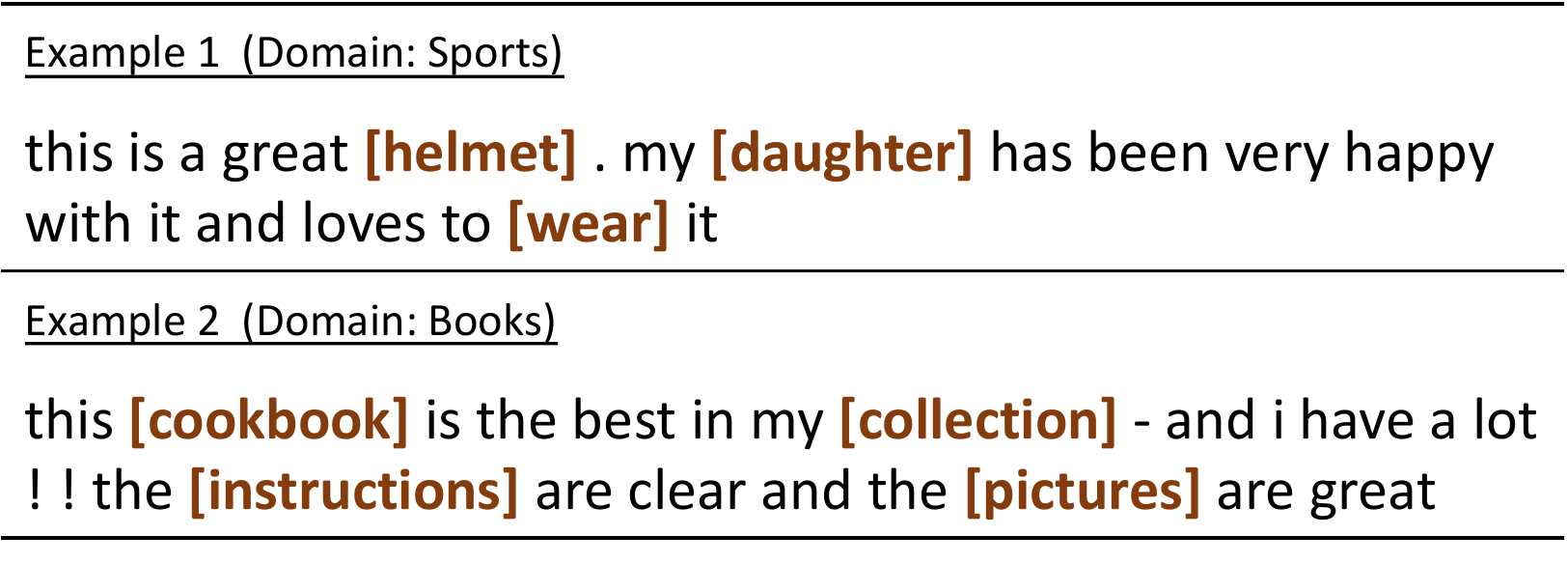}
	\caption{Two examples from Sport and Books domains that illustrate our motivation for transforming sentences to be \textit{domain-invarint}  by masking  domain-related words in square brackets.}
	\label{fig:motivation}
\end{figure}
	
% current methods
In this work, we focus on the task of multi-domain sentiment classification (MDSC) where we need to make full use of limited annotated data and large unlabeled data from each domain to train a classifier that achieves best average performance on all domains. There exist two major lines of work attempting to tackle this challenge. One line is to exploit the shared-private framework \cite{NIPS2016_6254,wu2015collaborative}, where domain-agnostic features are captured by the networks shared across all domains and domain-specific representations by feature extractor of each domain.  \cite{liu-etal-2017-adversarial,chen-cardie-2018-multinomial} applied domain adversaries to shared features for better learning of domain-invariant representations. The other major line of work \cite{liu-etal-2018-learning,ijcai2018-642,ijcai2019-681} implicitly utilized such share-private ideas where they first learned domain-specific query vectors (or domain embeddings) and then used these to compose domain-aware representation by attending features from shared sentence encoder. So far, these two major methods have not been effectively combined.

\begin{figure*}
	\centering
	\includegraphics[width=5in]{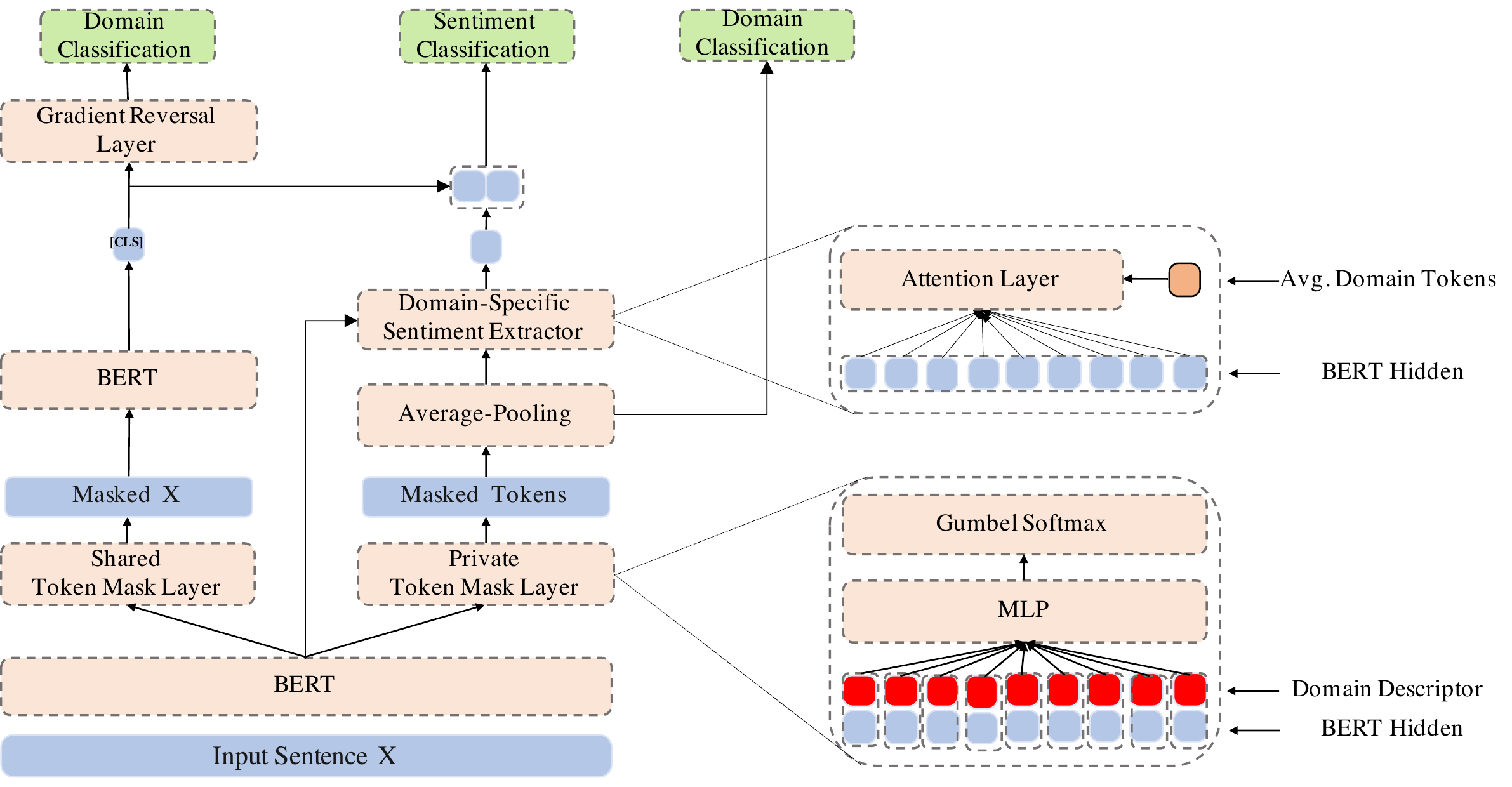}
	\caption{An overall architecture of BertMasker.}
	\label{fig:model_bert}
\end{figure*}

While shared-private models learn domain-agnostic features in high-dimensional space,  the discrimination between shared and private features cannot be directly interpreted to humans at text-level. Therefore, we propose to distinguish domain-related and domain-agnostic tokens before further feature extractions, based on the intuition that removing domain-related words from texts would help improve their domain-invariance.  Given two sentences from Sports and Book domains respectively in Figure \ref{fig:motivation}, after removing domain-related words like \textit{helmet} from Sports domain and \textit{cookbook} from Book domain, these sentences become more domain-agnostic. Meanwhile, the most salient sentiment-related semantics are mostly preserved in the remaining texts. In this way, it would be possible to tell what features are domain-related and what features are shared by all domains to some extent.  

To combine the advantage of both paradigms in multi-domain sentiment analysis, a model should employ the shared-private framework, where the shared part learns domain-agnostic sentiment features and the private part captures a domain-aware sentiment representation based on the shared feature extractors (contrary to separate extractors for each domain in \cite{liu-etal-2017-adversarial}). To learn  good shared sentiment features with better interpretablity at text level, a model should be capable of discriminating between domain-related and domain agnostic tokens at first. To this end, we propose the BertMasker model. The BertMasker model learns to first select domain-related tokens from texts, then mask those tokens from original text and acquires domain-agnostic sentiment features for the shared part. As the masked tokens are domain-related, they are appropriate for learning domain-aware sentiment representations of texts from different domains. We incorporate this advantage into the private part of BertMakser. Since simple models are not adequate for learning good sentiment features from fractional texts, we turn to BERT \cite{devlin-etal-2019-bert} for text encoding as it shares similar input format during its pre-training phase of Masked Language Model (MLM). Motivated by previous work \cite{sun-etal-2019-utilizing} utilizing Next Sentence Prediction task in BERT, we also expect inputting texts with \textit{MASK} at both training and inference time would boost the performance in multi-domain sentiment analysis tasks. Though we have no accurate prior knowledge of what domain-related tokens are, the BertMasker takes a detour of learning domain-related tokens as we have some knowledge of what domain-related tokens should not be for our sentiment classification task. In other words, tokens from general sentiment lexicons and commonly used stopwords are domain-agnostic. We enhance this prior knowledge as constraints to our model and train a domain classification model to guide more accurate learning of domain-related tokens. Those tokens serves as the key to learn both shared and private sentiment features.

Our contributions can be summarized as follows:
\begin{itemize}
	\item We propose a novel model named BertMasker to better learn shared representation across domains by masking domain-related tokens from text.
	\item Our model combines both shared-private framework and domain-aware feature learning, where token masking networks in the shared part of learns domain-invariant text transformation and in the private part to aggregates domain-aware sentiment features.
	\item Evaluation results on benchmark multi-domain sentiment classification datasets demonstrate the superiority of our proposed model.  Further analysis on masked tokens and remaining texts proves the plausibility and effectiveness of token masking mechanism.
\end{itemize}

\section{Related Work}

Our work uses a BERT-based model for multi-domain sentiment classification. We describe related work from these two perspectives.

\subsection{Multi-domain Sentiment Classification}
The task of multi-domain sentiment classification \cite{li-zong-2008-multi} aims at training models that leverage data from multiple domains to improve the overall classification performance on all domains. Currently, there exists mainly two lines of related methods.  One line of methods \cite{liu-etal-2017-adversarial,liu-etal-2018-learning,chen-cardie-2018-multinomial,NIPS2016_6254} is to exploit shared-private framework, where domain-agnostic features are usually captured with adversarial training or gradient reversal layer \cite{ganin2016domain} at the shared part. Meanwhile, domain-specific representations are learnt by feature extractors of each domain. Further, \cite{guo-etal-2018-multi,chen-etal-2019-multi-source} apply mixture-of-expert \cite{Jacobs:1991:AML:1351011.1351018} approach to explicitly capture knowledge shared among similar domains. The other line of methods \cite{ijcai2018-642,ijcai2019-681} is to learn domain representations through domain classification and use these as queries to acquire domain-sensitive representations of input texts. 

Our proposed BertMasker combines the power of both paradigms. It employs a shared-private framework. It first learns to select domain-related words. Then, our model obtains shared sentiment features by exploiting texts without those words and using these selected words to obtain domain-aware sentiment representations. 

\subsection{BERT-based Models in Sentiment Analysis}

BERT is one of the key innovations in the recent advances of contextualized representation learning \cite{mccann2017learned,peters-etal-2018-deep,howard-ruder-2018-universal,devlin-etal-2019-bert}. Its success relies on two pre-training tasks, namely Masked Language Model (MLM) and Next Sentence Prediction (NSP). Currently, there are mainly two ways to utilize BERT for downstream tasks. One is fine-tuning for each end task. For instance, to make the input format consistent with that of NSP, \cite{sun-etal-2019-utilizing} constructs auxiliary sentences for aspect-based sentiment classification in four ways. And the other is injecting task-specific knowledge \cite{ke2019sentilr,tian-etal-2020-skep} using new pre-training tasks. Such injections are usually done along with MLM where other objectives like POS tag, sentiment polarity \cite{ke2019sentilr}, and sentiment targets \cite{tian-etal-2020-skep} are introduced.

Our work is partially motivated by \cite{sun-etal-2019-utilizing}, as we both transform inputs to have same format as one of the pre-training task of BERT (they use NSP while we use MLM instead). To our best knowledge, we are the first to make usage of MLM at both training and test time in sentiment analysis tasks. As MLM aims at predicting masked words based on context from both left and right, the BERT model can recover the semantic of current word being masked. In other words, the BERT model could retain much of overall features of the sentences while a small portion of its constituent tokens being masked. We make use of such advantage and design a model that could automatically learn to mask some (domain-related) words. In this way, a sentence could be transformed to be domain-invariant while still retaining its most salient sentiment features. 

\section{Model}

\subsection{Overview}

Basically, our model adopts the popular adversarial shared-private framework, where the shared part is utilized for extracting domain-invariant features and private part for domain-specific features. An overview of our model is shown in Figure \ref{fig:model_bert}. For a given sentence, BertMasker first models each word in its context. Then it determines whether a token/word is  domain-related with token masking  networks from shared and private part. In the shared part, it replaces those domain-related tokens with \textit{[MASK]} symbol in original text  and feed the transformed, more domain-invariant text to BERT for invariant sentiment feature learning. In the private part, it instead choose those \textit{masked} tokens to learning domain-aware sentiment representations with attention mechanism. Finally, the concatenation of shared and private features are used for sentiment prediction. In the following, we introduce key components of our model in details.

\subsection{Sentence Modeling with BERT}
% BERT as token encoder
Given an input sequence $X= \{x_1, x_2, \ldots, x_N\}$, we first get the contextualized representation $h_i$ of tokens $x_i$ using BERT encoder:
\begin{equation}
\{h_1, h_2, \ldots, h_N\} = BERT(x_1, x_2, \ldots, x_N) \\
\label{equ:bert_model}
\end{equation}

The MLM task of BERT enables it processing sequences whose tokens are partially replaced with \textit{[MASK]} symbol. We exploit such intrinsic advantage of BERT to facilitate our idea of modeling text after removal of domain-related tokens. Suppose $K$ tokens in the given sequence are selected and we have the \textit{masked} text $X^M= \{x_1, x_2, \ldots, [Mask]_1, \ldots, [Mask]_K, \ldots,  x_N\}$. Then we could model the new text with Equation \ref{equ:bert_model} and obtain $H^{masked} = \{h_1, h_2, \ldots, h_{{mask}_1}, \ldots, h_{{mask}_K}, \ldots,    h_N\}$.  

Following previous sentiment classification tasks using BERT, we can choose the hidden feature $h_{cls}$ from token \textit{[CLS]} as sequence representation.  

\subsection{Token Masking Networks}
In this part, we describe how to select domain-related words and generate masked results for shared and private part.

\paragraph{Shared Part}

It is straightforward to assume that if we remove tokens that are domain-related from a text sequence, the remaining part should be more domain-agnostic than the original sequence. 
Based on this assumption, the token masking network first decides whether a token $x_i$  should be selected  according to its contextualized representation $h_i$.  Follow the idea  in \cite{liu-etal-2018-learning}, we also introduce domain descriptors $D=\{d_1,, d_2, \ldots, d_j, \ldots, d_{M}\}$ for each domain to encode most representative characteristics of that domain from the training data, where $M$ is the number of domains involved in training and test.  As we have the domain label for every instance at both training and test time, we can use its domain descriptors to help decide whether a token is highly correlated with the current domain. For each token $x_i$,  we combine its hidden representation $h_i$ and its domain descriptor $d_j$ as $z_i$.  We use simple feed-forward neural works with \textit{tanh} non-linearity (referred to as \textit{MLP}) for measuring relatedness $l_i$ between a token and the domain of its text. To generate discrete decision of mask, we apply GumbelSoftmax \cite{jang2016categorical} to $l_i$. The choice $P_{shared}$ of removing token $x_i$ or not can be calculated as follows:
\begin{eqnarray}
\label{equ:mask_combine}
z_i &=& h_i \oplus d_j   \\
\label{equ:mask_mlp}
l_i &=& MLP(z_i) \\
\label{equ:mask_gumbel}
p_i &=& GumbelSoftmax(l_i) 
\end{eqnarray}

\paragraph{Private Part}

The main difference between token masking networks in shared and private part is that 

Instead of only using domain descriptor $d_j$ of current text,  we adopt a mixture of domain descriptors for each sequence in private part. The motivation behind this is to better collect domain-related words for each individual sentence if it shares similarities with other domains. We treat $h_{CLS}$ as the current sentence representation and measure its relatedness to all domain descriptors. We apply simple inner-product attention through following equations:
\begin{eqnarray}
z_j &=& h_{CLS} \oplus d_j   \\
\hat{z_j} &=& MLP(z_j)  \\
s_i & = &  \langle d_i, \hat{z_j} \rangle \\
a_i & = & Softmax(s_i) \\
\hat{d_j} &=& \sum_{1}^{M}a_i * d_i
\end{eqnarray}
where $d_j$ is the corresponding domain descriptor of current text and $M$ is the number of domains. Then we follow similar steps of Equation \ref{equ:mask_combine}-\ref{equ:mask_gumbel} except that $d_j$ in Equation \ref{equ:mask_combine} is replaced with $\hat{d_j}$.
We denote the masking result of private  part as $P_{private}$.

\paragraph{Masking Constraints}
Though final sentiment classification result is good for BertMasker without preset rules, preliminary results show that masking rate is  over 30\%  on average for all domains and many irrelevant tokens are selected, making the remaining texts less interpretable for humans. This may owe to the existence of non-robust features \cite{NIPS2019_8307} in these multi-domain datasets.
Thus, to alleviate this, we add some human priors of what we think should not be domain-related words, namely stop words and sentiment words in lexicons. We explicitly ignore relevant masking decisions.  We use stop words from this site \footnote{https://github.com/amueller/word\_cloud} and sentiment words from 
\cite{hu2004mining}. We also add common negation and intensifier words into the constraints.

\subsection{Domain-invariant Sentiment Feature Extraction}
After acquiring the masking result $P_{shared}$ from token masking network in the shared part, we replace the chosen words with \textit{[MASK]} symbol and feed the new sequence into the shared BERT again. We use hidden output $\hat{h_{cls}^s}$ of token \textit{[CLS]} as domain-invariant sentiment representation.

\paragraph{Adversarial Feature Learning}

To ensure sequence representations learned from previous masking process are domain-agnostic, we pass the feature outputs $h_{cls}^s$ from BERT through a Gradient Reversal Layer \cite{ganin2016domain} and perform domain classification as follows:
\begin{eqnarray}
h_{grl} &=& GRL(h_{cls}^s) \\
% l_d &=& W_4Tanh(W_3h_{grl} + b_3) + b_4 \\
\label{equ:dc_mlp}
l_d &=& MLP(h_{grl}) \\
p_d &=& Softmax(l_d)  \\
\label{equ:dc_loss}
L_{ds} &=& CrossEntropy(p_d, y_d)
\end{eqnarray}

Here, we adopt cross-entropy loss and we refer to $h_{cls}^s$  as $h_{shared}$.

\subsection{Domain-specific Sentiment Feature Extraction}

\paragraph{Domain Informative Feature}

Similarly, we can obtain domain-related tokens $X_j = \{x_{j_1}, x_{j_2}, \ldots, x_{j_K}\}$ from private token mask layer, where $K$ is the number of selected domain-related tokens in input sequence $P_{private}$. Then,  hidden representation of those tokens is aggregated as domain-related clue $h_j$:
\begin{equation}
h_j = \frac{1}{K} \sum_{j_1}^{j_K}{h_{j_t}}
\end{equation}
Besides, we enforce these clues to be domain-related following Equation \ref{equ:dc_mlp}-\ref{equ:dc_loss} and the loss is $L_{dp}$.

\paragraph{Domain-aware Sequence Encoding}

Since we have the domain-related clue $h_j$, we can use it as the query vector and apply attention mechanism to find the most relevant feature of current input and its corresponding domain. Here, we use simple inner-product attention for its simplicity. The formulas are listed as follows:
\begin{eqnarray}
\beta_t &=& \langle h_j, h_t \rangle \\
\alpha_t &=& Softmax(\beta_t) \\
h_{private} &=& \sum_{1}^{N}\alpha_t * h_t
\end{eqnarray}

\subsection{Sentiment Classification}
The final feature for sentiment classification is the concatenation of $h_{shared}$ and $h_{private}$. We use a shared sentiment classifier for all domains and the probability of each sentiment is calculated as follows:
\begin{eqnarray}
\label{equ:concat}
h_{concat} &=& h_{shared} \oplus h_{private} \\
\label{equ:sc_mlp}
l_s &=& MLP(h_{concat} ) \\
\label{equ:sc_softmax}
p_s &=& Softmax(l_s)
\end{eqnarray}
We use the cross entropy loss between the predictions and true labels as $L_s$ for training a sentiment classifier:
\begin{equation}
\label{equ:sc_loss}
L_s = CrossEntropy(p_s, y_s)
\end{equation}

To make feature representations from shared and private part more sentiment-oriented and further boost the performance for final sentiment classification, we add layers similar to Equation \ref{equ:sc_mlp}-\ref{equ:sc_loss} to $h_{shared}$ and $h_{private}$ and get their corresponding loss as $L_{ss}$ and $L_{sp}$.

\subsection{Final Loss}
The total loss of our model can be computed as follows:
\begin{equation}
\resizebox{1\hsize}{!}{$L_{all} = \lambda_{ds} * L_{ds} + \lambda_{dp} * L_{dp} + \gamma * L_s + \gamma_{ss} * L_{ss} + \gamma_{sp} * L_{sp} + \lambda \left \| \theta \right \|^2$}
\end{equation}
where $\lambda_{ds}$ and $\lambda_{dp}$ are coefficients for domain classification, $\gamma$, $\gamma_{ss}$, and $\gamma_{sp}$ are coefficients for sentiment classification.

\section{Experiments}

\begin{table}
	\centering
	\begin{adjustbox}{center}
		\begin{tabular}{l | c c c c }
			\hline
			Dataset & Train & Dev. & Test & Avg. Length \\
			\hline
			Books  & 1400 & 200 & 400 & 159 \\
			Electronics & 1398 & 200 & 400 & 101 \\
			DVD & 1400 & 200 & 400 & 173 \\
			Kitchen & 1400 & 200 & 400 & 89 \\
			Apparel & 1400 & 200 & 400 & 57 \\
			Camera & 1397 & 200 & 400 & 130 \\
			Health & 1400 & 200 & 400 & 81 \\
			Music & 1400 & 200 & 400 & 136 \\
			Toys & 1400 & 200 & 400 & 90 \\
			Video & 1400 & 200 & 400 & 156 \\
			Baby & 1300 & 200 & 400 & 104 \\
			Magazines & 1370 & 200 & 400 & 117 \\
			Software & 1315 & 200 & 400 & 129 \\
			Sports & 1400 & 200 & 400 & 94 \\
			IMDB & 1400 & 200 & 400 & 269 \\
			MR & 1400 & 200 & 400 & 21 \\
			\hline
		\end{tabular}
	\end{adjustbox}
	\caption{Statistics of  datasets from 16 domains.}
	\label{table:dataset}
\end{table}

\begin{table*}
	\small
	\centering
	\begin{adjustbox}{width=\columnwidth*2,center}
		\begin{tabular}{ l | c c c  | c c c c c c}
			\hline
			\multirow{2}{*}{\bf Domain} & \multicolumn{3}{c|}{\bf Single Domain}  & \multicolumn{6}{c}{\bf Multiple Domains} \\
			& {\bf BLSTM} & {\bf CNN} & {\bf BERT} & {\bf ASP-MTL }  & {\bf DA-MTL } & {\bf DSR-at } & {\bf DAEA } & {\bf DAEA+BERT}   & {\bf BertMasker }\\
			\hline\hline
			Books & 81.00 & 85.30 & 87.00 & 84.00 & 88.50 & 89.10 & 89.00 & N/A & \textbf{93.00} \\
			Electronics & 81.80 & 87.80 & 88.30 & 86.80 & 89.00 & 87.90 & 91.80 & N/A & \textbf{93.25} \\
			DVD & 83.30 & 76.30 & 85.60 & 85.50 & 88.00 & 88.10 & 88.30 & N/A & \textbf{89.25} \\
			Kitchen & 80.80 & 84.50 & \textbf{91.00} & 86.20 & 89.00 & 85.90 & 90.30 & N/A & 90.75 \\
			Apparel & 87.50 & 86.30 & 90.00 & 87.00 & 88.80 & 87.80 & 89.00 & N/A & \textbf{92.25} \\
			Camera & 87.00 & 89.00 & 90.00 & 89.20 & 91.80 & 90.00 & 92.00 & N/A & \textbf{92.75} \\
			Health & 87.00 & 87.50 & 88.30 & 88.20 & 90.30 & 92.90 & 89.80 & N/A & \textbf{95.25} \\
			Music & 81.80 & 81.50 & 86.80 & 82.50 & 85.00 & 84.10 & 88.00 & N/A & \textbf{89.50} \\
			Toys & 81.50 & 87.00 & 91.30 & 88.00 & 89.50 & 85.90 & 91.80 & N/A & \textbf{93.75} \\
			Video & 83.00 & 82.30 & 88.00 & 84.50 & 89.50 & 90.30 & \textbf{92.30} & N/A & 91.25 \\
			Baby & 86.30 & 82.50 & 91.50 & 88.20 & 90.50 & 91.70 & 92.30 & N/A & \textbf{92.75} \\
			Magazines & 92.00 & 86.80 & 92.80 & 92.20 & 92.00 & 92.10 & \textbf{96.50} & N/A & 94.50 \\
			Software & 84.50 & 87.50 & 89.30 & 87.20 & 90.80 & 87.00 & 92.80 & N/A & \textbf{93.00} \\
			Sports & 86.00 & 85.30 & 90.80 & 85.70 & 89.80 & 85.80 & 90.80 & N/A & \textbf{92.50} \\
			IMDB & 82.50 & 83.30 & 85.80 & 85.50 & 89.80 & \textbf{93.80} & 90.80 & N/A & 86.00 \\
			MR & 74.80 & 79.00 & 74.00 & 76.70 & 75.50 & 73.30 & 77.00 & N/A & \textbf{83.75} \\
			\hline
			Avg & 83.80 & 84.49 & 88.16 & 86.09 & 88.61 & 87.86 & 90.16 & 90.5 & \textbf{91.47} \\
			\hline
		\end{tabular}
	\end{adjustbox}
	\caption{Results of multi-domain sentiment classification. Accuracy (\%) is adopted for evaluation.}
	\label{tabel:multi_classification}
\end{table*}

\subsection{Dataset}
We use the dataset from \cite{liu-etal-2017-adversarial} \footnote{http://pfliu.com/paper/adv-mtl.html} for multi-domain sentiment classification task. which consists product and movie reviews from 16 domains. Following previous work, we partition dataset in each domain into training, development and testing sets according to the proportion of 70\%, 10\%, and 20\%. The detailed statistics of all the datasets are listed in Table \ref{table:dataset}.

\subsection{Implementation Details}
We adopt BERT$_{base}$, to be specific, its implementation \footnote{https://github.com/huggingface/transformers} in PyTorch for all the experiments. The maximum sequence length for BERT model is set to 128. The mini-batch size is set to 8 and we train the model for 15 epochs. The model with highest averaged accuracy on development sets is chosen for final comparison. SGD is applied to optimize all our models with an initial learning rate of 0.0003.  The coefficients $\lambda_{ds}$ and $\lambda_{dp}$ for domain classification loss are set to 0.002 and $\gamma$, $\gamma_{ss}$, and $\gamma_{sp}$ for sentiment classification loss are chosen as 0.4, 0.3, 0.3 respectively.  The dimension is set to 200 for domain descriptors. For multi-domain sentiment classification, we train over domain classification task for the first 2000 steps and sentiment classification in each domain for the next 3000 steps. After that, we train the model with both sentiment classification loss and domain classification loss. For cross-domain experiments, the only difference is that no target sentiment data is used during the latter two training phase.

\begin{table}
	\begin{adjustbox}{width=\columnwidth,center}
		\centering
		\begin{tabular}{l | c c c c }
			\hline
			& ASP-MTL & DSR-at & DAEA & BertMasker \\
			\hline
			Books & 81.50 & 85.80 & 87.30 & \textbf{87.75} \\
			Electronics & 83.80 & 89.50 & 85.80 & \textbf{93.00} \\
			DVD & 84.50 & 86.30 & \textbf{88.80} & 87.75 \\
			Kitchen & 87.50 & \textbf{88.30} & 88.00 & 87.76 \\
			Apparel & 85.30 & 85.80 & 88.00 & \textbf{91.25} \\
			Camera & 85.30 & 88.80 & 90.00 & \textbf{91.50} \\
			Health & 86.00 & 90.50 & 91.00 & \textbf{94.75} \\
			Music & 81.30 & 84.80 & 86.50 & \textbf{89.50} \\
			Toys & 88.00 & 90.30 & 90.30 & \textbf{91.50} \\
			Video & 86.80 & 85.30 & \textbf{91.30} & 89.00 \\
			Baby & 86.50 & 84.80 & 90.30 & \textbf{91.50} \\
			Magazines & 87.00 & 84.00 & 88.50 & \textbf{91.25} \\
			Software & 87.00 & \textbf{90.80} & 89.80 & 90.50 \\
			Sports & 87.00 & 87.00 & 90.50 & \textbf{91.25} \\
			IMDB & 84.00 & 83.30 & 85.80 & \textbf{86.75} \\
			MR & 72.00 & 76.30 & 75.50 & \textbf{82.50} \\
			\hline
			Avg & 84.59 & 86.35 & 87.96 & \textbf{89.84} \\
			\hline
		\end{tabular}
	\end{adjustbox}
	\caption{Results of cross-domain (15-to-1) sentiment classification. Accuracy (\%) is adopted for evaluation.}
	\label{table:cross_classification}
\end{table}

\subsection{Multi-domain Classification}

We experiment with multi-domain sentiment classification on 16 test sets respectively. We compare with several baselines and previous state-of-the-art models.

\textbf{Single Task}. We use a bi-directional LSTM and a simple CNN model as single task baselines which are trained on each domain independently.

\textbf{BERT}. BERT \cite{devlin-etal-2019-bert} is a pre-trained contextualized representation learning model which has achieved state-of-the-art results on many tasks. We use pre-trained BERT-base model and fine-tune it for each domain.

\textbf{ASP-MTL}. The model used in \cite{liu-etal-2017-adversarial} with adversarial training on the shared part and separate LSTMs for each domain in the private part.

\textbf{DA-MTL}. DA-MTL \cite{ijcai2018-642} dynamicly generates query vector for each instance and then uses this query vector to attend over the hidden representations of input sentence.

\textbf{DSR-at}. DSR-at \cite{liu-etal-2018-learning} is also based on share-private scheme. Different from ASP-MTL, it applies memory network as private feature extractor.

\textbf{DAEA}. DAEA \cite{ijcai2019-681} is an attention based method which first generates domain-specific query vector and domain-aware word embeddings. It then use the query vector to attend over the hidden representations from BLSTM with domain-aware word embeddings as input.

\textbf{DAEA+BERT}. DAEA+BERT \cite{ijcai2019-681} improves DAEA by using BERT as word initialization. It is the previous state-of-the-art model in multi-domain sentiment classification.

We present results of multi-domain text classification in Tabel \ref{tabel:multi_classification}. Generally, using data from multiple domains improves average classification performance. We can see that BERT achieves superior performance on single domain setting, even outperforming ASP-MTL and  DSR-at which simultaneously use sentiment classification data from multiple domains. Our model BertMasker achieves the best performance, outperforming other methods in 12 out of  16 domains. 

Compared with the state-of-the-art DAEA+BERT model, our model still achieve 0.93\% performance gain in terms of  average accuracy of all domains. It is worth noting that our model brings nearly 6.75\% increase in MR domain.The main reason for the improvement may lie in that we utilize both shared and private part while DAEA+BERT model only exploits domain-aware private representations. Besides, this also demonstrates some shared sentiment features shared across domains may not be well captured by a domain-aware feature extractor. Therefore, such performance gain quantifies the BertMasker's capability in learning good sentiment features shared by all domains.

Our model \textit{fails} by a large margin compared to other multi-domain models on IMDB domain, one possible reason maybe the used maximum length for BERT is much smaller than average sentence length in IMDB (128 compared to 256). For domains like Magazines, as single BLSTM along could achieve comparable performances, the shared part contributes little to final sentiment prediction. Thus, our model is outperformed by DAEA+BERT model which is skilled at learning domain-aware sentiment representations.

\subsection{Cross-domain Experiments}

Unlike multi-domain sentiment classification,  the task of cross-domain sentiment classification doesn't provide any training data for a target domain. Thus, it calls for better utilization of the shared knowledge across all domains. To further understand whether BertMasker achieves such capability, we also test our model on 15-to-1 cross-domain sentiment classification setting \cite{liu-etal-2017-adversarial,ijcai2019-681}, where models are trained using the training data of sentiment and domain classification from 15 domains and training data of domain classification from the target domain.

As shown in Table \ref{table:cross_classification}, our model achieves 1.88\% performance gain in averaged accuracy of all domains compared to previous best performing model DAEA. Besides, it outperforms all the other  models in 12 out of 16 domains on cross-domain sentiment classification task. These results confirm the superiority of masking networks in BertMasker, which manifests in learning better shared representations for sentiment classification than other models.

\subsection{Ablation Test}
\begin{table}
	\centering
	\begin{tabular}{l | c }
		\hline
		& Avg. Accuracy (\%) \\
		\hline
		\textit{w/o} shared part & 91.0 \\
		\textit{w/o} private part & 88.56  \\
		\hline
		\textit{w/o} shared mask & 90.69  \\
		\textit{w/o} private mask & 91.21 \\
		\textit{w/o} sentiment word mask & 91.29 \\
		\textit{w/o} stop word mask & 91.17 \\
		\hline
		Full model & 91.47 \\
		\hline
	\end{tabular}
	
	\caption{Ablation test results of BertMasker on multi-domain sentiment classification. Average accuracy is presented. \textit{w/o} stands for \textit{without}.}
	\label{table:ablation}
\end{table}

To further explore how well each component contributes to the prediction of sentiment, we carry out an ablation study  of BertMasker on test set in multi-domain setting.  As shown in Table \ref{table:ablation},  the performance decreases when removing either the shared or private network and the removal of  private part leads to more performance loss compared the shared part.  An intuitive explanation is that the domain-aware sentiment features integrate both domain-agnostic and domain-specific sentiment features through attention mechanism. Moreover, token masking networks in shared and private part help increase the performance by 0.78\% and 0.26\% respectively, which prove the effectiveness of our proposed masking method. 

\begin{figure*}
	\begin{subfigure}{.5\textwidth}
		\centering
		\includegraphics[width=.8\linewidth]{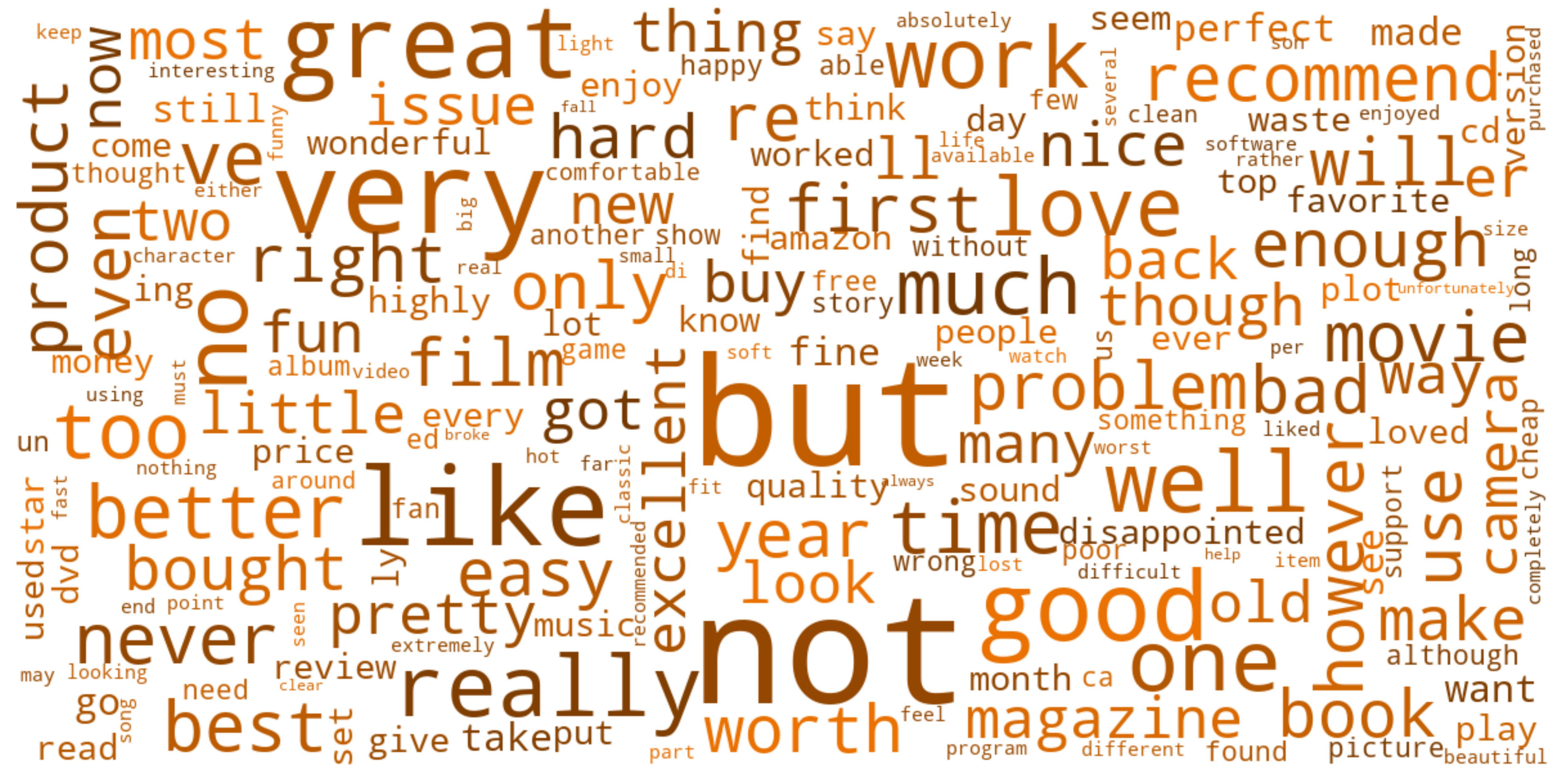}
		\caption{All domains: Top domain-invariant words}
		\label{fig:wc_dc_mask}
	\end{subfigure}
	\begin{subfigure}{.5\textwidth}
		\centering
		\includegraphics[width=.8\linewidth]{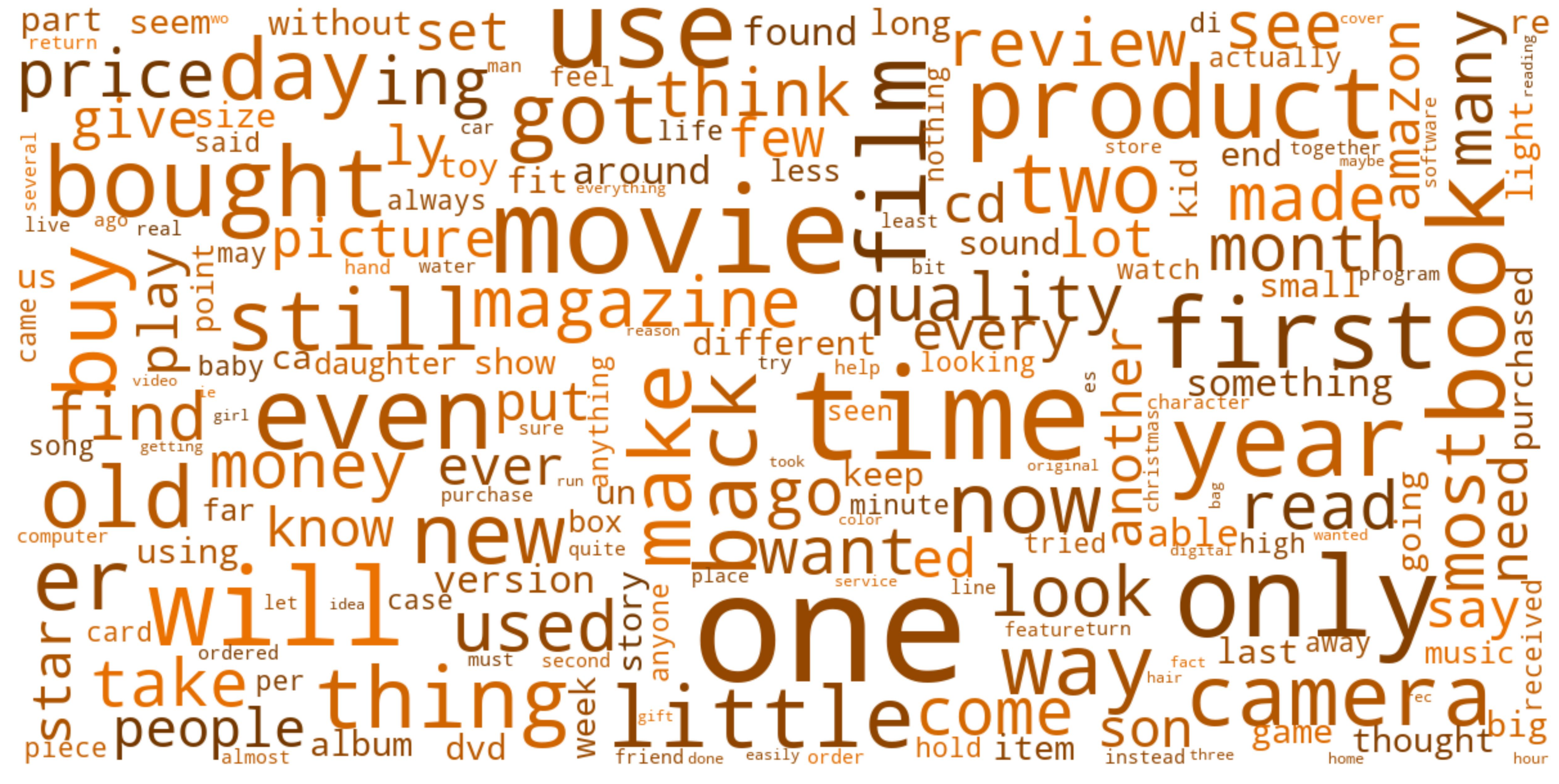}
		\caption{All domains: Top domain-related words}
		\label{fig:wc_sc_mask}
	\end{subfigure}
	\begin{subfigure}{.5\textwidth}
		\centering
		\includegraphics[width=.8\linewidth]{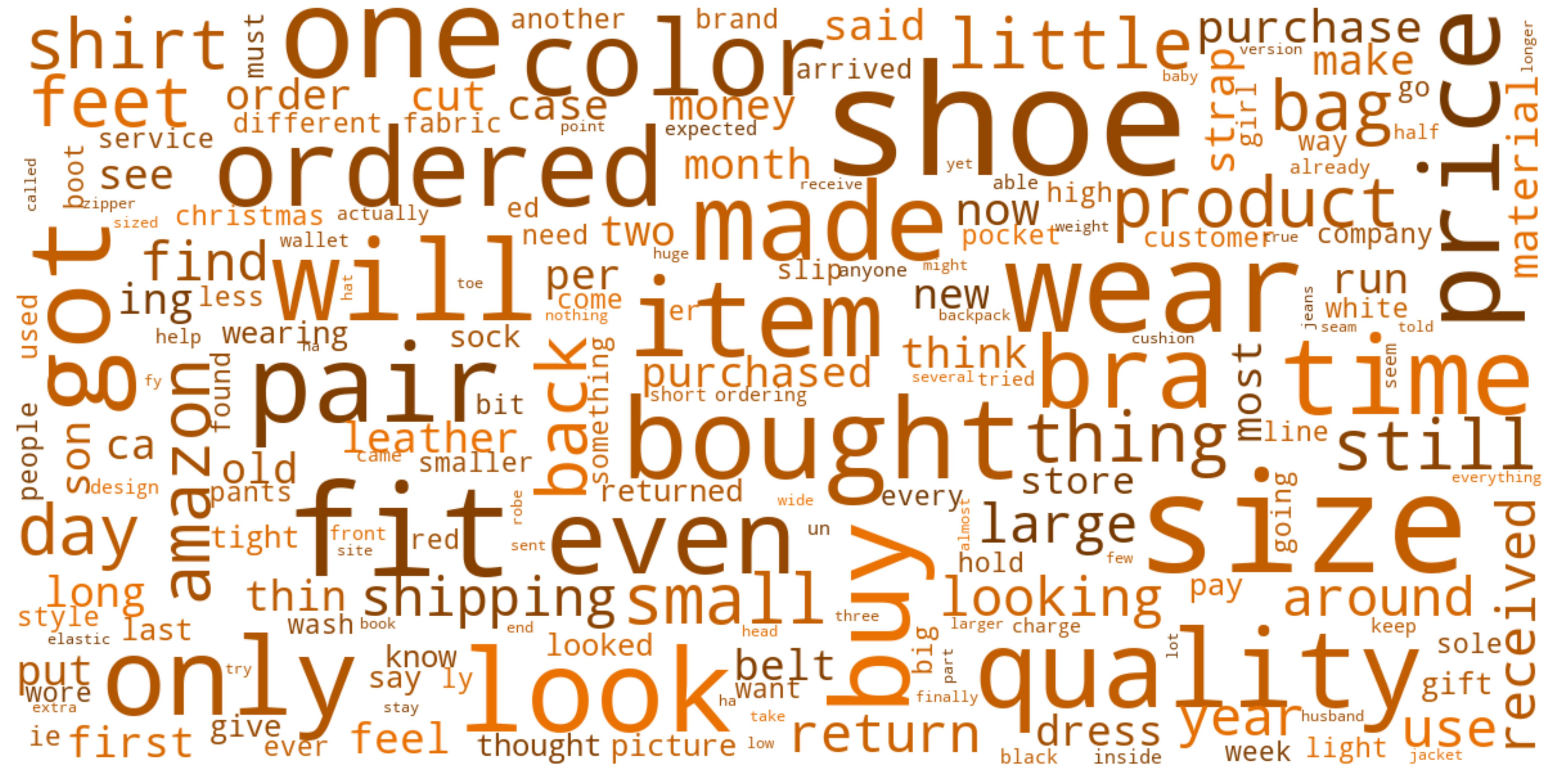}
		\caption{Apparel: Top domain-related words}
		\label{fig:wc_apparel_mask}
	\end{subfigure}
	\begin{subfigure}{.5\textwidth}
		\centering
		\includegraphics[width=.8\linewidth]{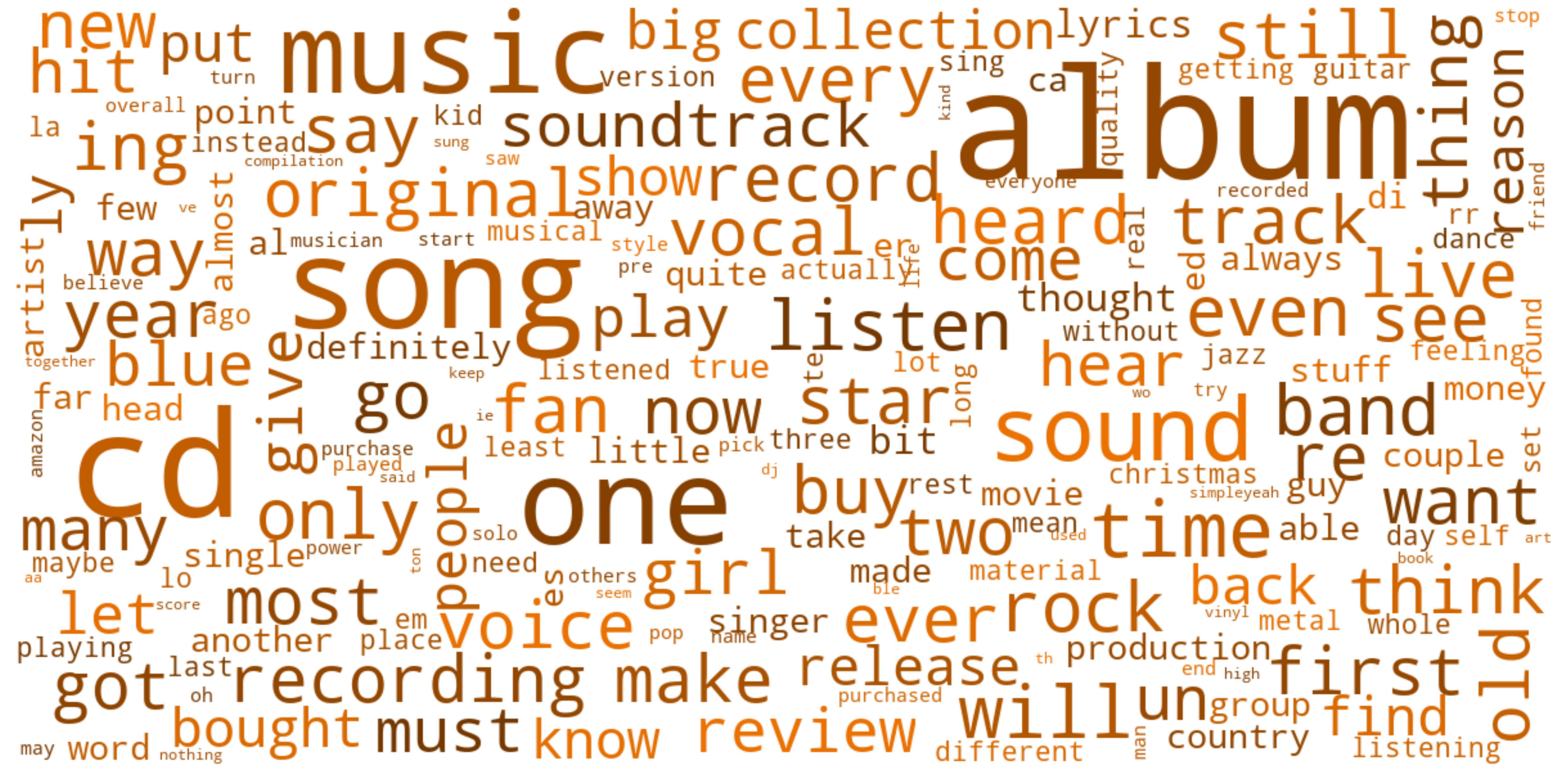}
		\caption{Music: Top domain-related words}
		\label{fig:wc_music_mask}
	\end{subfigure}
	\caption{Word cloud of Tokens from Token Masker Layer. Larger word size means higher frequency of occurrence.}
	\label{fig:wc_mask}
\end{figure*}

\textbf{Masking constraints} Besides, further experiments on two masking constraints demonstrate both stop words masking and sentiment words masking improve the performance of BertMasker, by 0.30\% and 0.16\% respectively.

\section{Analysis of Masking}

In this part, we conduct several quantitative and qualitative experiments with BertMasker on it masking part.
\begin{table}
	\begin{adjustbox}{width=\columnwidth,center}
		\centering
		\begin{tabular}{l | c | c | c  }
			\hline
			& Shared (No./Portion) & Private (No./Portion) & Avg. Length \\
			\hline
			Books & 25.60/0.13 & 23.59/0.12 & 190 \\
			Electronics & 21.76/0.17 & 20.14/0.16 & 128 \\
			DVD & 23.78/0.10 & 22.68/0.10 & 226 \\
			Kitchen & 21.21/0.19 & 22.20/0.20 & 111 \\
			Apparel & 15.09/0.20 & 15.25/0.21 & 74 \\
			Camera & 23.18/0.16 & 24.48/0.16 & 148 \\
			Health & 19.86/0.20 & 18.87/0.19 & 101 \\
			Music & 22.38/0.14 & 21.69/0.13 & 162 \\
			Toys & 22.04/0.20 & 21.18/0.19 & 112 \\
			Video & 23.05/0.12 & 20.33/0.11 & 191 \\
			Baby & 23.91/0.19 & 21.64/0.17 & 128 \\
			Magazines & 22.88/0.16 & 20.45/0.14 & 144 \\
			Software & 21.14/0.14 & 21.74/0.14 & 151 \\
			Sports & 21.28/0.17 & 21.10/0.17 & 125 \\
			IMDB & 29.17/0.11 & 27.53/0.10 & 264 \\
			MR & 6.34/0.23 & 5.85/0.21 & 27 \\
			\hline
			Avg. & 21.42/0.15 & 20.54/0.14 & 143 \\
			\hline
		\end{tabular}
	\end{adjustbox}
	\caption{The number and percentage of masked words in shared and private part of BertMasker on test set in multi-domain sentiment classification setting.}
	\label{table:mask_portion}
\end{table}

% \subsubsection{Number of words masked}
\subsection{Number of Words Masked} 

As observed in Table \ref{table:mask_portion}, the number and percentage of masked tokens of each domain correlates with its average sentence length, where in general domains with longer average sequence length have more tokens masked and lower masking rate. Another interesting finding is that the final masking rates of both shared and private part are similar to the percentage (15\%) of \textit{[MASK]} token in the Mask Language Model pre-training task of BERT. We leave it as a future work to explore whether this rate correlates with implementation of mask in BERT or the number of domain-related tokens in original data distribution.

%\subsubsection{Top words masked}
\subsection{Top Words Masked}  

Apart from the number and percentage of masking, we also would like to investigate whether the token masking networks of BertMasker mask meaningful words. In Figure \ref{fig:wc_dc_mask} and \ref{fig:wc_sc_mask}, we use word cloud to illustrate tokens after masking from shared part and masked tokens from private part of all domains.  Besides, we exhibit masked tokens from private part of Apparel and Music domains. From Figure \ref{fig:wc_sc_mask} we can see that, the top tokens from the masked sequence in shared part are mostly domain-invariant sentiment-related words, which includes polarity words like \textit{good, great, well}, negation words like \textit{but, not, no} and intensifiers like \textit{ really, very}.  This demonstrates that after token masking network removing domain-related tokens, the shared part focus more on domain-invariant sentiment features. From Figure \ref{fig:wc_apparel_mask} and \ref{fig:wc_music_mask}, we find that as two domains share less opinion targets, the distribution of domain-related tokens from their corresponding private masking networks are quite different from each other, where Apparel domain can be depicted with words like \textit{fit, shoes, wear, size, shirt, .etc} and Music domain can be represented using words including \textit{album, song, sound, music, cd, .etc}.  When analyzing Figure \ref{fig:wc_dc_mask}, we find that no domain-related words obviously outnumber the other words in the private part from all domains, which again shows the distinction of data distribution of domains in the datasets.

\subsection{Domain Classification After Masking} 

\begin{table}
	\begin{adjustbox}{center}
		\centering
		\begin{tabular}{l | c  }
			\hline
			&  Accuracy (\%)\\
			\hline
			On masked sequences & 65.21 \\
			\hline
			On original sequences & 77.95 \\
			\hline
			On masked words & 65.67 \\
			\hline
		\end{tabular}
	\end{adjustbox}
	\caption{Results of domain classification on original sequences, sequences after removing masked words and masked words.}
	\label{table:domain_mask}
\end{table}

To further verify whether masking ``domain-related'' tokens from a text improves its domain-invariance, we conduct domain classifications on both original and masked texts. Here, we utilize BERT-base as a powerful feature extractor and apply an MLP similar to \ref{equ:dc_mlp} for domain classification. We evaluate the results using accuracy.

As shown in Table \ref{table:domain_mask}, it's relatively easy to distinguish domains based on original texts. Our mask network successfully degrades the domain classification performance by over 11\% on masked texts. This reveals that our strategy of masking is working towards our expectations of domain-invariant text. However, as we don't have direct knowledge of what domain-related tokens are, tokens extracted using the masking network constrained by external sentiment and stopword lexicons are sub-optimal for domain classification task. Thus, the result demonstrates that the remaining text still contains abundant clues for domain classification.

\begin{figure}
	\begin{subfigure}{.5\textwidth}
		\centering
		\includegraphics[width=.6\linewidth]{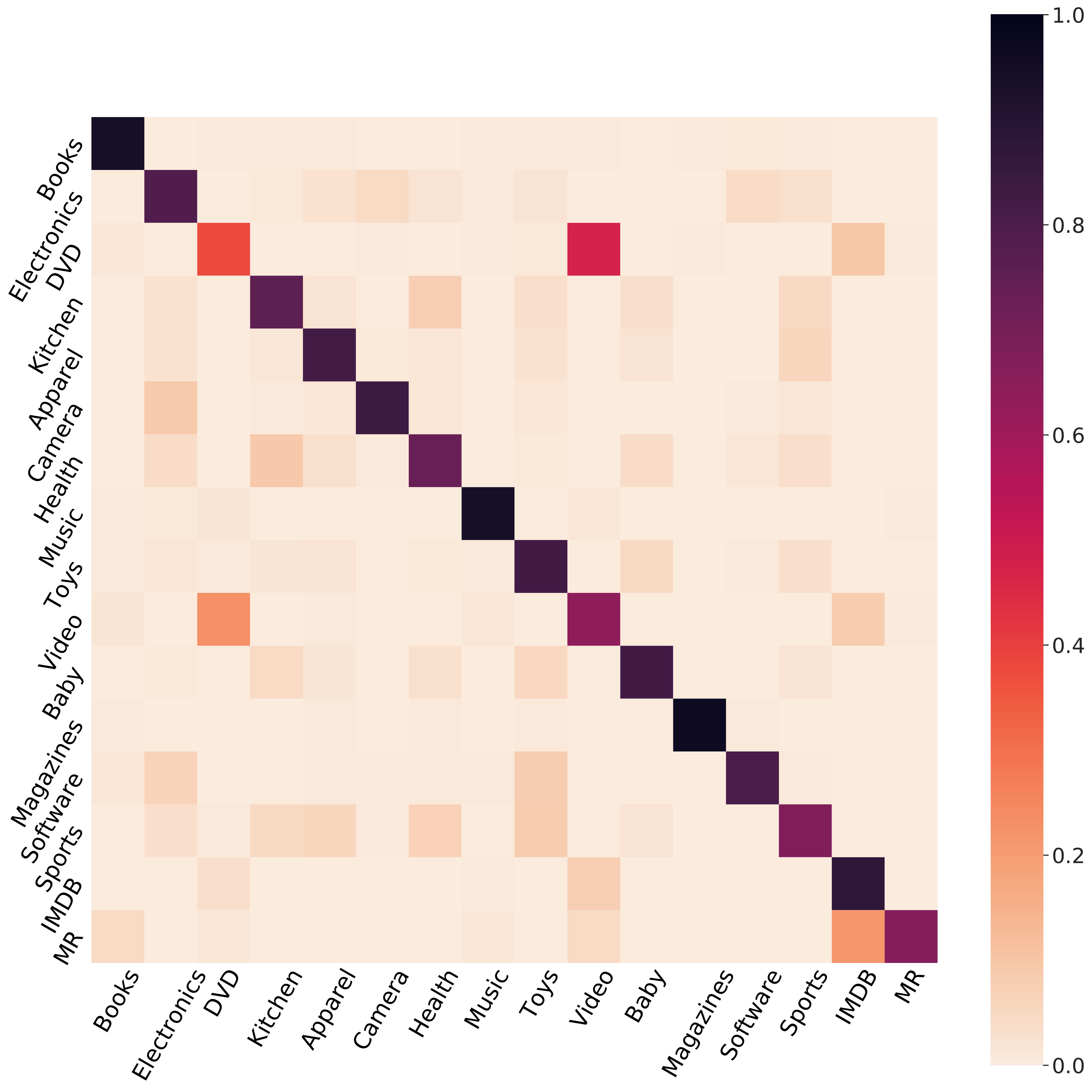}
		\caption{On original sequences}
		\label{fig:cm_original}
	\end{subfigure}
	\begin{subfigure}{.5\textwidth}
		\centering
		\includegraphics[width=.6\linewidth]{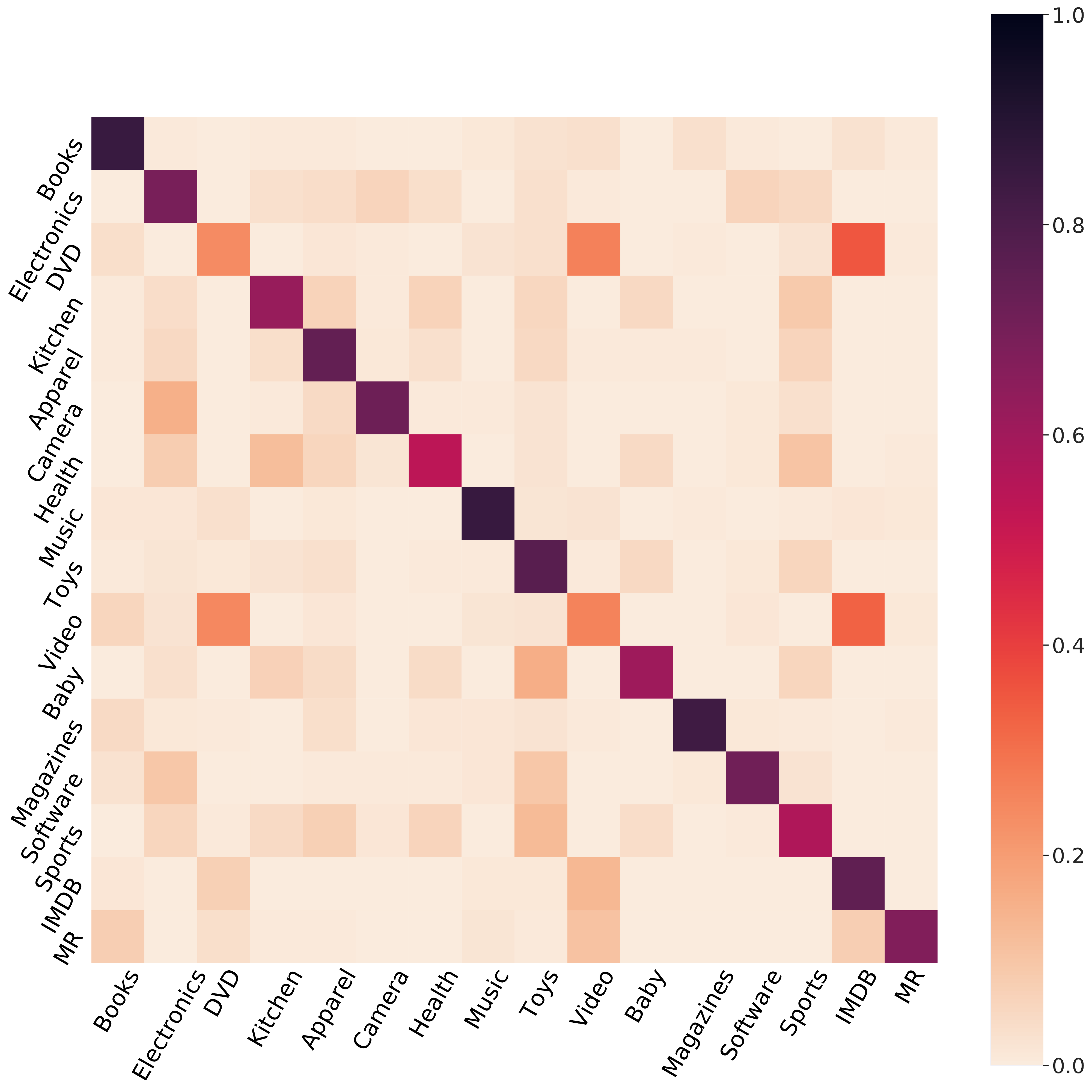}
		\caption{On masked sequences}
		\label{fig:cm_masked}
	\end{subfigure}
	
	\caption{Confusion matrices of domain classification on original and masked sequences.}
	\label{fig:cm_mask}
\end{figure}

To further explore how the masking works on each domain, we visualize the confusion matrices of domain classification on original and remaining text separately in Figure \ref{fig:cm_mask}. For example, by comparing the \textit{Sports} row in Figure \ref{fig:cm_original} and Figure \ref{fig:cm_masked}, we can see that shallow blocks in \ref{fig:cm_original} become darker in \ref{fig:cm_masked} and opposite case happens to darker blocks. This reflects that domain classifier can't find necessary features on the remaining texts, thus mis-classifies more cases into domains sharing some similarities with \textit{Sports} domain, eg. \textit{Electronics}, \textit{Toys}, \textit{Camera} and even on \textit{Software} domain.

From the above experiments, we can see that token-level masking strategy works to help transform the sentence to be more domain-invariant in the shared part and selecting domain-related words for better domain-aware sentiment feature learning.

\subsection{Case Study and Error Analysis}
\begin{figure}
	\centering
	\includegraphics[width=2.9in]{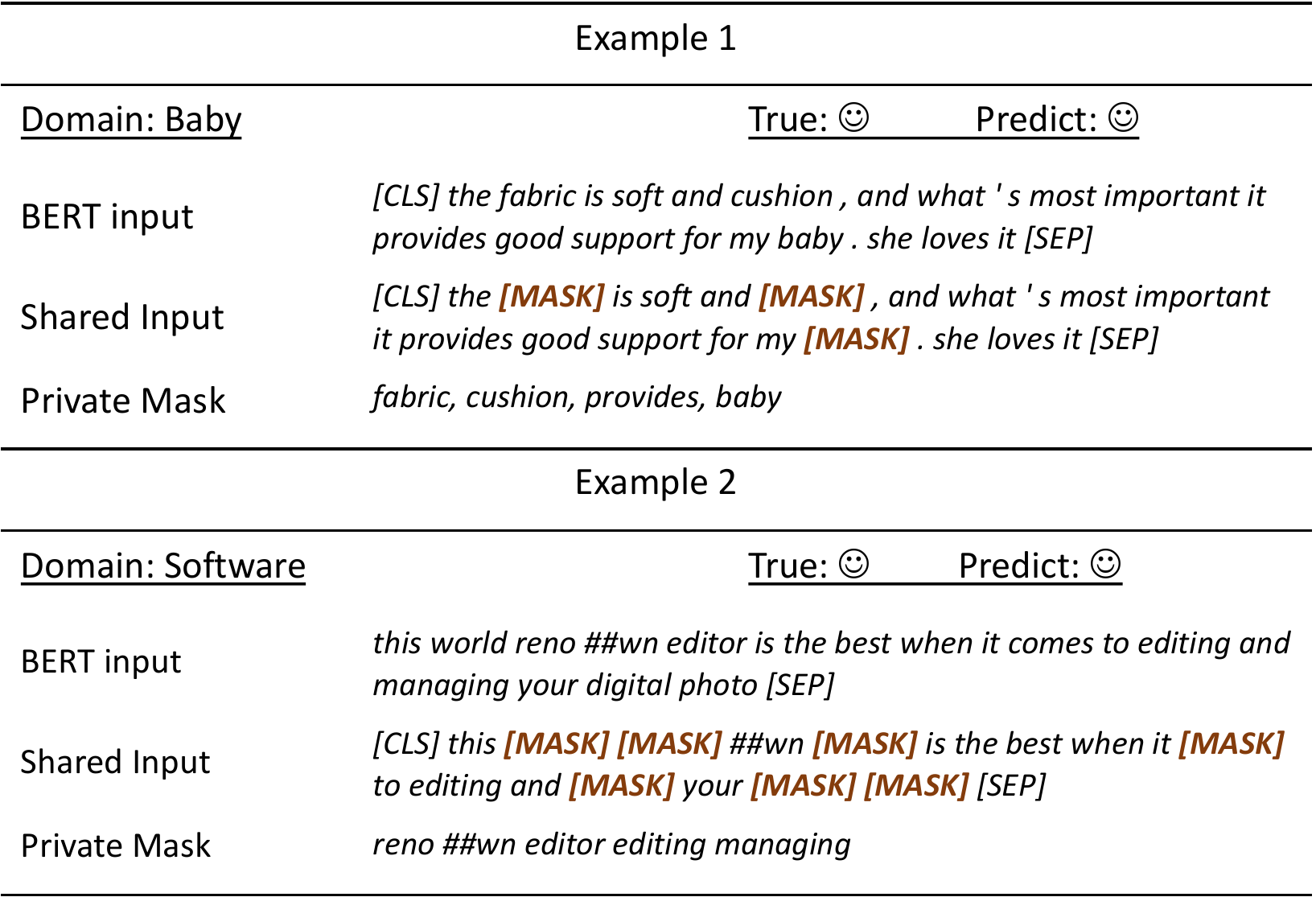}
	\caption{Visualization of masked words in two sentences from Magazine and Baby domains.}
	\label{fig:mask_example}
\end{figure}

We visualize the words selected by token masking networks in BertMasker from both shared and private part  in Figure \ref{fig:mask_example}. As illustrated in Example 1, the model successfully masks domain-related words like \textit{fabric, cushion, baby} in the sentence and make correct sentiment predictions based on those both domain-invariant and domain-aware representations. However, we also notice in many cases, due to the existence of unknown words and errors incurred by  word-piece tokenizer used by BERT, the masked tokens may not be semantically adequate or meaningful. From Example2, we can see that as \textit{renown} is not recognized by BERT, it further influences the masking result in the shared part. Besides, as we employ different masking strategies in shared and private part, the masking results diverge in both examples.

\section{Conclusion}

In this paper, we propose the BertMasker model which combines the power of two popular paradigms in multi-domain sentiment classification: shared-private structure and shared encoder with domain-aware aggregation. Instead of directly learning domain-variant features in a high-dimensional space, we propose to first transform sentences to be more domain-invariant through \textit{masking} domain-related words, which alongside utilizes the power of BERT in learning good semantic representations from \textit{masked} texts. Our model outperforms existing works in both multi-domain and cross-domain settings on the benchmark dataset. Detailed analysis of the \textit{masked} words further proves effectiveness of our proposed \textit{masking} strategy.

In the future, we would like to work on two directions: (1) replace the mask network with simpler network, e.g. distilled BEERT models, to accelerate training and inference of our model. (2) incorporate more external knowledge to guide fine-grained and accurate selection of domain-related words and phrases.

\bibliography{main}
\bibliographystyle{acl_natbib}

\end{document}